\pdfoutput=1
\documentclass[11pt]{article}

\usepackage[]{EACL2023}

\usepackage{times}
\usepackage{latexsym}

\usepackage[T5]{fontenc}
\usepackage[utf8]{inputenc}

\usepackage{microtype}

\usepackage{inconsolata}

\usepackage{longtable}
\usepackage{graphicx,graphics}
\usepackage{tabularx}
\usepackage{arydshln}
\usepackage{float}
\usepackage{multirow}
\usepackage{pbox}
\usepackage{array,xcolor}
\usepackage{amsmath}
\usepackage{url}
\usepackage{pgfplots}
\usepackage{footnote}
\usepackage[flushleft]{threeparttable}
\usepackage{supertabular}
\usepackage{makecell}
\usepackage{multicol}
\usepackage{xcolor, soulutf8}
\definecolor{RED}{RGB}{237,125,49}
\sethlcolor{RED}

\title{ViHOS: Hate Speech Spans Detection for Vietnamese}

\author{Phu Gia Hoang, 
  Canh Duc Luu,
  Khanh Quoc Tran, \\
  {\bf Kiet Van Nguyen}, 
  {\bf Ngan Luu-Thuy Nguyen}\\
  University of Information Technology, Ho Chi Minh City, Vietnam \\
  Vietnam National University, Ho Chi Minh City, Vietnam \\
  \texttt{\{19520215, 19521272\}@gm.uit.edu.vn} \\
  \texttt{\{khanhtq, kietnv, ngannlt\}@uit.edu.vn}
  }

\begin{document}
\maketitle
\begin{abstract}
The rise in hateful and offensive language directed at other users is one of the adverse side effects of the increased use of social networking platforms. This could make it difficult for human moderators to review tagged comments filtered by classification systems. To help address this issue, we present the ViHOS (\textbf{Vi}etnamese \textbf{H}ate and \textbf{O}ffensive \textbf{S}pans) dataset, the first human-annotated corpus containing 26k spans on 11k comments. We also provide definitions of hateful and offensive spans in Vietnamese comments as well as detailed annotation guidelines. Besides, we conduct experiments with various state-of-the-art models. Specifically, XLM-R$_{Large}$ achieved the best F1-scores in Single span detection and All spans detection, while PhoBERT$_{Large}$ obtained the highest in Multiple spans detection. Finally, our error analysis demonstrates the difficulties in detecting specific types of spans in our data for future research. Our dataset is released on GitHub\footnote{\url{https://github.com/phusroyal/ViHOS}}.

\textbf{Disclaimer}: This paper contains real comments that could be considered profane, offensive, or abusive. 
\end{abstract}

\section{Introduction}
\label{introduction}
Social networking sites have been widely used all over the world. Here, users can easily share their thoughts, connect with others, or earn money by selling items, creating content, and so on. Since these sites are universally accepted, many extreme users misuse comment functions to abuse other individuals or parties with hate and offensive language. Consequently, it has been proved that these types of speech could harm other users’ health (\citealp{mohan2017impact}; \citealp{anjum2018empirical}). Sometimes these behaviors can be considered cyberbullying, cyber threats, or online harassment.

\begin{figure}[]
\centering
\includegraphics[width=1\linewidth]{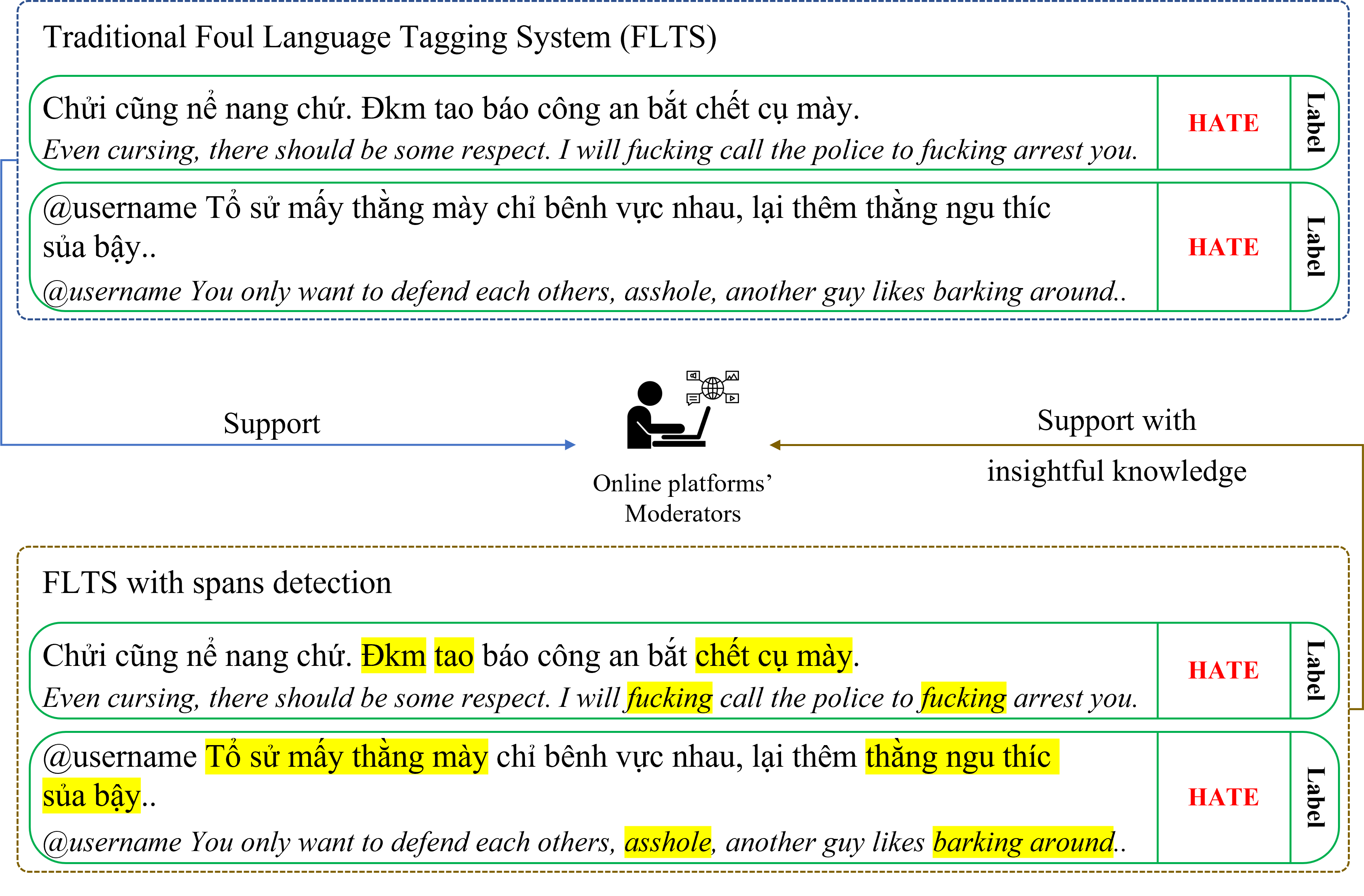} 
\caption{An example of the aid of spans detection for traditional foul language tagging system can provide additional insightful knowledge about tagged comments for human moderators.}
\label{fig:guideline}
\end{figure}

However, current studies are mainly about classifying comments as a whole with binary labels (\citealp{zampieri-etal-2019-predicting}; \citealp{nguyen2021constructive}) or multiple labeling schemes of abusive behaviors (\citealp{davidson2017automated}; \citealp{founta2018large}; \citealp{mathur-etal-2018-detecting}). These efforts are made to aid human moderators, who need to review a massive number of online tagged comments that violate their community standards. However, a system that can highlight the spans that make a comment hateful or offensive can be more advantageous to human moderators who frequently deal with long and tedious comments and prefer explanations over a system-generated unexplained tag per comment. Furthermore, in some cases, using highlighted spans and moderators' context knowledge, they can take some actions to stop cyberbullying or online harassment. Nevertheless, there is only a study on toxic spans, SemEval-2021 Task 5: Toxic Spans Detection \cite{pavlopoulos2021semeval}. On the other hand, in a study of \citet{mathew2020hatexplain}, hate and offensive spans worked as a rationale to support models in classifying the whole comments.

In Vietnamese, the resources about hate and offensive language are limited, namely ViHSD \cite{luu2021large}, HSD-VLSP \cite{vu2020hsd}, and UIT-ViCTSD \cite{nguyen2021constructive}. Indeed, there is no study about hate and offensive spans in Vietnamese. This motivated us to (i) develop a new task of extracting hate and offensive spans from Vietnamese social media texts that conceivably impact research and downstream applications and (ii) provide the Natural Language Processing (NLP) research community with a new dataset for recognizing hate and offensive spans in Vietnamese social media texts. 

Our two main contributions are summarized:

\begin{enumerate}
  \item We created the first human-annotated dataset for Vietnamese Hate and Offensive Spans (ViHOS) comprising 26,467 human-annotated spans on 11,056 comments. Our dataset is annotated with a clear definition of hate and offensive spans, along with detailed and specialized guidelines for a less-studied language like Vietnamese. Compared to the toxic spans dataset at SemEval-2021 Task 5 \cite{pavlopoulos2021semeval}, which is built to detect toxic spans from toxic comments, or the HateXplain dataset \cite{mathew2020hatexplain}, which has spans working as a rationale for classifying the whole sentence, ours includes not only a large number of texts with annotated hate and offensive spans but also clean texts without any spans. This effort is made to serve a new task of detecting hate and offensive spans from Vietnamese online social media comments.

  \item To evaluate the efficacy of our dataset, strong baselines are empirically investigated on ViHOS, including BiLSTM-CRF \cite{DBLP:journals/corr/LampleBSKD16}, XLM-R \cite{conneau2019unsupervised}, and PhoBERT \cite{nguyen2020phobert}. We conducted various experiment schemas, including comparing the full dataset having additional clean comments with the dataset that does not have; Single span detection, Multiple spans detection, and All spans detection. We obtain that: (i) Additional clean comments help the baselines have better performance than the dataset without them for 10$\pm$2\% (ii) After fine-tuning the deep learning model and pre-trained language models, results show that the pre-trained language models outperform the deep learning models.

\end{enumerate}
\section{Related work}
\label{relatedworks}
To the best of our knowledge, much of the research in the field of hate speech detection has been conducted in English due to the abundance of corpora and the robust pre-trained models. Many benchmark datasets for hate and offensive speech in other languages have also been published in recent years, including Arabic \cite{mubarak2020arabic}, Dutch \cite{tulkens2016dictionary}, and French \cite{chiril2019multilingual}. Novel models are introduced to improve the efficiency of hate and offensive speech detection. Initial approaches were based on typical machine learning and deep neural networks with word embeddings. Transformer-based models such as BERT \cite{devlin2018bert}, BERTology \cite{rogers2020primer}, and BERT-based transfer learning \cite{ruder2019transfer} have recently been used to detect hate and offensiveness that achieved competitive results in major SemEval shared tasks such as SemEval-2020 Task 12 \cite{zampieri2020semeval}, and SemEval-2021 Task 5 \cite{pavlopoulos2021semeval}. However, research in Vietnamese is still limited in terms of the dataset and experimental methods. Only a few outstanding research exist, such as ViHSD \cite{luu2021large}, HSD-VLSP \cite{vu2020hsd}, and UIT-ViCTSD \cite{nguyen2021constructive}.

For the topic of detecting foul spans, there are only a few case studies in English that are closely related, namely the SemEval-2021 Task 5: Toxic Spans Detection dataset \cite{pavlopoulos2021semeval} and the HateXplain dataset \cite{mathew2020hatexplain}. The toxic spans, defined in the SemEval-2021 Task 5 dataset, are the sequences of words that make a text toxic. There are a total of 10,629 posts in this dataset, which stems from the Civil Comments dataset \cite{borkan2019nuanced}. Another dataset with hate and offensive spans at the word level is HateXplain. The HateXplain contains 20,148 Gab and Twitter posts. Each post is manually classified into one of three labels: hateful, offensive, and normal. %The annotators in this study were also required to highlight the words that are important for classification.

In this study, we focus on Vietnamese to close the gap and \textbf{develop the first Vietnamese hate and offensive spans detecting benchmark}. 

%UIT-ViSD4SA \cite{nguyen2021span}, VLSP 2016 and 2018 NER shared task \cite{nguyen2016vlsp,nguyen2018vlsp}, and PhoNER\_COVID19 \cite{truong2021covid}. 

\section{Dataset Creation}
\label{dataset}
\subsection{Dataset Source}
\label{datasource}

\begin{figure*}[]
\centering
\includegraphics[width=1\linewidth, height=6cm]{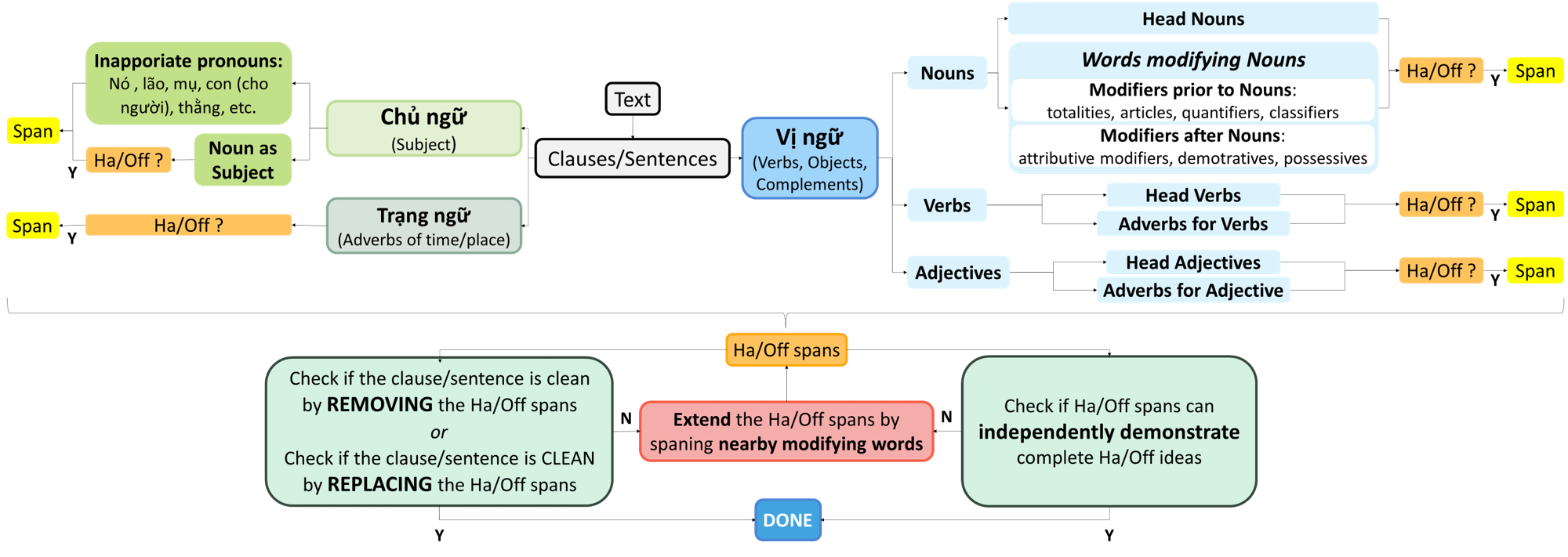} 
\caption{Detailed annotation guidelines for annotating comments for annotators. In which, \textit{Ha/Off?} stands for a requirement for annotators to check whether the associated components are hate (Ha) or offensive (Off) or not.}
\label{fig:guideline}
\end{figure*}

ViHOS consists of 11,056 comments derived from the ViHSD dataset \cite{luu2021large}. The Vietnamese Hate Speech Detection dataset (ViHSD) is one of the few large and credible social media text datasets in a low-resource language like Vietnamese. ViHSD contains 27,624, 3,514, and 2,262 of CLEAN, HATE, and OFFENSIVE comments, respectively. Comments in ViHSD are public and collected from social media platforms. Thus, metaphors, idioms, proverbs, and other tricky characteristics of online comments abound.

All of the HATE, OFFENSIVE comments from ViHSD after removing duplicates (5,528 comments left) are used to annotate the hate and offensive spans. Otherwise, 5,528 CLEAN comments, which also come from ViHSD and do not violate any hate or offensive definition defined in Section \ref{guidelines}, are manually annotated for our dataset. We append the 5,528 CLEAN comments because: (*) We aim to detect the hate and offensive spans directly in online comments; (**) With an equal number of span and non-span (clean) comments helps models not be biased towards any type.

\subsection{Annotation Guidelines}
\label{guidelines}

Our goal is to create a dataset that contains comprehensive hate and offensive thoughts, meanings, or opinions within the comments rather than just a lexicon of hate and offensive terms. We define the hate or offensive spans as follows to help annotators understand our goals:

\begin{itemize}
  \item Harassing, cursing, insulting, disrespecting others.
  \item Sexual or verbal abuse towards one or a group of individuals based on their sensitive characteristics such as region, religion, politics, body, gender, etc. 
  \item Insinuations, metaphors, metonymy used for hate, offensive or controversial purposes on sensitive issues such as region, gender, religion, politics, human rights, etc. 
  \item Disuniting any factions or parties based on their politics, religion, ideologies, genders, etc. 
  \item Causing verbal disrespect by using inappropriate pronouns. 
  \item If replaced or removed, the sentence will no longer be hateful or offensive. 
\end{itemize}

However, the hate or offensiveness in Vietnamese comments might cover one or even many components of a sentence. For instance: "thằng ad thở ra cái tư duy như trẻ lớp mầm" (\textbf{Eng:} \textit{the admin speaks as his mind just like a kindergarten boy.)} This comment consists of three nouns/nouns phrases: "thằng ad"\textsubscript{\textit{the admin}} (it is offensive when calling a guy as a "thằng") as subject; "cái tư duy"\textsubscript{\textit{mind}} (the appearance of the word "cái" causes this noun phase become offensive) and "trẻ lớp mầm"\textsubscript{\textit{kindergarten  boy}} as objects; one verb: "thở" (it usually means \textit{breathe}, but in this context, we could consider it as \textit{speak} but in a hate manner). As defined above, we must annotate a part of the subject, "thằng," and the whole phrase of the verb with its objects, "thở ra cái tư duy như trẻ lớp mầm" (\textbf{Eng:}\textit{ speaks as his mind just like a kindergarten boy}) in order to capture the whole hate or offensive ideas. 

\begin{table*}[h]
\centering
\begin{tabular}{l||l|l|l|l||c||c} 
\hline
\multirow{2}{*}{} & \multicolumn{4}{c||}{\textbf{Phase 1}}                                                                                                         & \multirow{2}{*}{\textbf{Phase 2 }} & \multirow{2}{*}{\textbf{Phase 3 }}  \\ 
\cline{2-5}
                  & \multicolumn{1}{c|}{\textbf{1$^{st}$}} & \multicolumn{1}{c|}{\textbf{2$^{nd}$}} & \multicolumn{1}{c|}{\textbf{3$^{rd}$}} & \multicolumn{1}{c||}{\textbf{4$^{th}$}} &                                    &                                     \\ 
\hline
Kappa score       & 0.4161                              & 0.4568                              & 0.4936                              & 0.6402                               & \multicolumn{1}{l||}{0.7239}         & \multicolumn{1}{l}{0.7215}     \\ 
\hdashline
F1-score          & 0.7085                              & 0.7186                              & 0.7534                              & 0.8219                               & \multicolumn{1}{l||}{0.8219}         & \multicolumn{1}{l}{0.8585}            \\
\hline
\end{tabular}
\caption{Inner-annotator agreement scores in three phases of annotation. In which, \textit{1$^{st}$, 2$^{nd}$, 3$^{rd}$, 4$^{th}$} are corresponding with four rounds of training annotators in Phase 1.}
\label{fig:iaa_scores}
\end{table*}

Therefore, we provided detailed guidelines (Figure \ref{fig:guideline}) to assist annotators in determining when to annotate one or multiple components in a hateful and offensive sentence. As we observed, most of the comments in our dataset are colloquial. These comments are written freely and lack many grammatical rules. As a result, we frequently witness comments that lack subject(s), verb(s), conjunctions, punctuation marks, and so on. To deal with this, the comments were split into clauses/sentences (units). The smallest subdivided unit must have at least a cluster of "Chủ ngữ"-"Vị ngữ" ("Subject"-"Verb, objects, complements"), which is considered as a simple sentence or clause, or a cluster of "Chủ ngữ"-"Vị ngữ" nesting in another "Chủ ngữ"-"Vị ngữ" as a complex sentence or clause. From this, annotators can recognize components in clauses or sentences before annotating them.

Furthermore, in Appendix \ref{apx:Abusive} and Appendix \ref{apx:notiecs}, we also provided notices for annotators in Table \ref{tab:ann_notices} and ways to help them deal with nine online foul linguistic phenomena in Table \ref{tab:cmt_Characteristic} while annotating the data. 

\subsection{Dataset Construction Process}

For dataset construction, we conducted three phases in which Phases 2 and 3 were inspired by \citet{truong-etal-2021-covid} and used metrics as bellow to calculate Inter-Annotator Agreement (IAA) among annotators. LightTag \cite{perry-2021-lighttag} is the tool we used for annotating data.

\textbf{Assessment of Inter-Annotator Agreement} \label{iaa}
Cohen's Kappa is widely used to measure inter-annotator agreement (IAA) in most tasks and is accepted as the standard measure \cite{mchugh2012interrater}. However, numerous studies indicated that Kappa is not the most proper measure for the NLP sequence tagging task like NER (\citealp{hripcsak2005agreement}; \citealp{grouin2011proposal}). The reason is that the definite number of negative cases required to calculate the Kappa does not exist for named entity spans. Spans in our task are the sequences of characters rather than sequences of tokens since hate and offensive spans could be icon(s), word(s), or distinct character set(s) (see Table \ref{tab:cmt_Characteristic} for more details). Therefore, the pre-existing fixed number of characters to consider in the process of annotating is not existent. 

A solution to deal with this is to calculate a character-level-based Kappa. Still, it has two associated problems: (1) annotators need to look at sequences of one or more characters instead of characters alone, causing the Kappa not to reflect the annotation task well; and (2) the "O"-labeled characters (the negative cases) outnumber the hate and offensive ones (the positive cases), provoking the Kappa to be computed on highly imbalanced data. For these reasons, the F1-score calculated without the negative cases is usually the measure for calculating IAA for the NLP tagging tasks like NER \cite{deleger2012building}. In this paper, IAA based on both F1-score (macro average) and character-level-based Kappa are calculated, while the former is the primary measure.

\begin{table*}[]
\centering
\begin{tabular}{l||r|r|r}
\hline
& \multicolumn{1}{c|}{\textbf{Train}} 
& \multicolumn{1}{c|}{\textbf{Dev}}  
& \multicolumn{1}{c}{\textbf{Test}}  \\ \hline
Number of clean comments            & 4,552           & 569   & 575   \\ \hdashline
Number  of Ha/Off comments          & 4,422           & 553   & 553   \\ \hline
Average clean comments length       & 8.69           & 8.50   & 9.04  \\ \hdashline
Average Ha/Off comments length      & 16.81          & 17.68 & 16.13 \\ \hline
Clean comments vocabulary size      & 4,234           & 1,423  & 1,400  \\ \hdashline
Ha/Off comments vocabulary size     & 5,162           & 2,089  & 2,013  \\ \hline
Number of multi-span comments (\%)  & 2,322 (26.26)          & 308 (27.85)   & 296 (26.76)   \\ \hdashline
Number of single-span comments (\%) & 1,970 (22.27)          & 229 (20.70)  & 235 (21.25)   \\ \hdashline
Number of non-span comments (\%)    & 4,552 (51.47)          & 569 (51.45)  & 575 (51.99)   \\ \hline
Average number of spans                & 2.10            & 2.09  & 2.00   \\ \hline
\end{tabular}
\caption{ViHOS statistics. Vocabularies size and comments length are calculated at the syllable level.}
\label{tab:data_stats}
\end{table*}

\textbf{Phase 1: Pilot Annotation}

Six undergraduate students were hired for our annotation tasks. The primary purpose of this pilot annotation phase was to familiarize our annotators with this task before entering the Main Annotation phase. We then developed an initial version of annotation guidelines with examples and distributed them to annotators. All annotators were required to carefully study the guidelines and give feedback before annotating the same 100 random samples from the 5,528 HATE, OFFENSIVE comments from ViHSD. This process was conducted four times with the F1-score and the Kappa for measuring IAA, which was calculated by averaging the results of pairwise comparisons across all annotators, shown in Table \ref{fig:iaa_scores}. All annotators were qualified as there was no F1-score of pairwise comparisons below 0.8.

\textbf{Phase 2: Ground Truth Annotation}

We randomly sampled a Ground Truth set of 600 comments from the 5,528 HATE, OFFENSIVE comments for this phase. Two guideline developers annotated the Ground Truth set separately using the well-developed guidelines from the former phase, resulting in an F1-score of 0.86 and Kappa (Cohen’s Kappa) of 0.72. Afterward, we hosted a discussion to deal with annotation conflicts and update the annotation guidelines.

\textbf{Phase 3: Main Annotation}

We split the remaining HATE and OFFENSIVE comments \cite{luu2021large} (4,928 comments left) into six non-overlapping and equal subsets. We also divided the 600-sample Ground Truth set from Phase 2 into six equal 100-sample smaller sets to insert into each subset. Each well-trained annotator from Phase 1 received a subset to annotate. Their annotation performance was assessed by calculating the F1 score and the Kappa score of the 100-sample Ground Truth sets in their subset. If any score is below 0.81 in terms of the F1 score, its corresponding annotator has to annotate again until it meets the requirement. This process was completed with an F1-score of 0.86 in the mean.

Furthermore, our annotators manually annotated CLEAN comments from ViHSD to spot any hate and offensive spans before being added to our dataset. This process collected 5,528 additional clean comments that met our requirements of having no hate and offensive spans.

\subsection{Dataset Statistics}

% \begin{table*}[]
% \centering
% \begin{tabular}{l||r|r|r}
% \hline
% & \multicolumn{1}{c|}{\textbf{Train}} 
% & \multicolumn{1}{c|}{\textbf{Dev}}  
% & \multicolumn{1}{c}{\textbf{Test}}  \\ \hline
% Number of clean comments            & 4,552           & 569   & 575   \\ \hdashline
% Number  of Ha/Off comments          & 4,422           & 553   & 553   \\ \hline
% Average clean comments length       & 8.69           & 8.50   & 9.04  \\ \hdashline
% Average Ha/Off comments length      & 16.81          & 17.68 & 16.13 \\ \hline
% Clean comments vocabulary size      & 4,234           & 1,423  & 1,400  \\ \hdashline
% Ha/Off comments vocabulary size     & 5,162           & 2,089  & 2,013  \\ \hline
% Number of multi-span comments (\%)  & 2,322 (26.26)          & 308 (27.85)   & 296 (26.76)   \\ \hdashline
% Number of single-span comments (\%) & 1,970 (22.27)          & 229 (20.70)  & 235 (21.25)   \\ \hdashline
% Number of non-span comments (\%)    & 4,552 (51.47)          & 569 (51.45)  & 575 (51.99)   \\ \hline
% Average spans length                & 2.10            & 2.09  & 2.00   \\ \hline
% \end{tabular}
% \caption{ViHOS statistics. Vocabularies size and comments length are calculated in syllable level.}
% \label{tab:data_statss}
% \end{table*}

Before conducting dataset analysis and experiments, ViHOS has a total of 11,056 comments after the annotation process and is divided into three subsets: train, development, and test, with an 8:1:1 ratio. In detail, ViHOS has 5,360 comments with hate and offensive spans and 5,696 clean comments without in which 5,528 comments were additionally added and 168 comments have no hate and offensive spans after Phase 3 in the annotation process. Table \ref{tab:data_stats} contains more information on the ViHOS statistics. It is apparent that the vocabulary of ViHOS is medium-sized, which is due to the small number of words in comments and comments in our dataset. In addition, more statistics about spans in ViHOS are shown in Table \ref{tab:spans_stats}.

\setbox0\hbox{\tabular{c}\textbf{Spans Quantity}\endtabular}
\setbox1\hbox{\tabular{c}\textbf{Spans Length}\endtabular}
\begin{table}[h]
\centering
\resizebox{\columnwidth}{!}{%
\begin{tabular}{cl||r|r|r} 
\hline
\multicolumn{1}{l}{}                                &                    & \multicolumn{1}{c|}{\textbf{Train}} & \multicolumn{1}{c|}{\textbf{Dev}} & \multicolumn{1}{c}{\textbf{Test}}  \\ 
\hline
\multirow{12.75}{*}{\rotatebox{90}{\rlap{\usebox0}}}& 0 span
  (\%)      & 4,552 (51.47)              & 569 (51.45)              & 575 (51.99)               \\ [1.25pt]
\cdashline{2-5}
                                                    & 1 span (\%)        & 1,970
  (22.27)            & 229 (20.71)              & 235 (21.25)               \\ [2.25pt]
\cdashline{2-5}
                                                    & 2 - 3 spans (\%)   & 1,527 (17.27)              & 207 (18.72)              & 202 (18.26)               \\ [2.25pt]
\cdashline{2-5}
                                                    & 4 - 6 spans (\%)   & 601 (6.80)                 & 75 (6.78)                & 68 (6.15)                 \\ [2.25pt]
\cdashline{2-5}
                                                    & 7 - 10 spans (\%)  & 164 (1.85)                & 18 (1.63)                 & 21 (1.90)                 \\ [2.25pt]
\cdashline{2-5}
                                                    &  >10 spans (\%)     & 30 (0.34)                 & 8 (0.72)                 & 5 (0.45)                                   \\  [2.25pt]
\hline
\multirow{10.85}{*}{\rotatebox{90}{\rlap{\usebox1}}}   & 1 syllable (\%)       & 5,253 (52.03)              & 699 (52.48)              & 647 (52.77)               \\ [2.25pt]
\cdashline{2-5}
                                                    & 2 - 3 syllables (\%)  & 3,554 (35.20)               & 466 (34.98)              & 474 (38.66)               \\  [2.25pt]
\cdashline{2-5}
                                                    & 4 - 6 syllables (\%)  & 916 (9.07)                & 122 (9.16)               & 112 (9.14)                \\ [2.25pt]
\cdashline{2-5}
                                                    & 7 - 10 syllables (\%) & 259 (2.57)                & 31 (2.33)                & 19 (1.55)                 \\ [2.25pt]
\cdashline{2-5}
                                                    &  >10 syllables (\%)    & 114 (1.13)                & 14 (1.05)                & 14 (1.14)                 \\ [2.25pt]
\hline
\end{tabular}%
}
\caption{Spans quantity and length statistics.}
\label{tab:spans_stats}
\end{table}

\begin{table*}[h]
\centering
\resizebox{\textwidth}{!}{
\begin{tabular}{l||c|c|c|c|c|c}
\hline
\textbf{} & \textbf{\begin{tabular}[c]{@{}c@{}}BiLSTM-CRF \\ + Pho2W$_{syllable}$\end{tabular}}& \textbf{\begin{tabular}[c]{@{}c@{}}BiLSTM-CRF \\ + Pho2W$_{word}$\end{tabular}} & \textbf{XLM-R$_{Base}$} & \textbf{XLM-R$_{Large}$} & \textbf{PhoBERT$_{Base}$} & \textbf{PhoBERT$_{Large}$} \\ \hline
\textbf{Full Data} & 0.7453 & 0.7036 & 0.7467 & \textbf{0.7770} & 0.7569 & 0.7716          \\ \hdashline \textbf{W/o additional clean comments} & 0.6241 & 0.6244 & 0.6479 & 0.6756          & 0.6738 & \textbf{0.6867} \\ \hline %Additional
\end{tabular}
}
\caption{Experimental results on Full Data versus Without additional clean comments.}
%\caption{Model performance results on the ViHOS test set}
\label{tab:results}
\end{table*}

% \begin{table}[h]
% \centering
% \begin{tabular}{lcc}
% \hline
% \textbf{Model}     & \textbf{F1-macro} & \textbf{F1-micro} \\ \hline
% BiLSTM-CRF         & 65.66             & 75.53             \\
% XLM-R              & 73.45             & 86.69             \\
% RoBERTa            & 72.07             & 86.08             \\
% \textbf{BERT case} & \textbf{74.05}    & \textbf{87.51}    \\
% BERT uncase        & 68.76             & 83.47             \\
% PhoBERT            & 73.88             & 87.47             \\ \hline
% \end{tabular}
% \caption{Model performance results.}
% \label{tab:results}
% \end{table}

\setbox2\hbox{\tabular{c}\textbf{Syllable}\endtabular}
\setbox3\hbox{\tabular{c}\textbf{Word}\endtabular}
\begin{table*}[]
\centering
\resizebox{\textwidth}{!}{
\begin{tabular}{ll||ccc|ccc|ccc}
\hline
\multirow{2}{*}{}  & \multirow{2}{*}{\textbf{Model}} & \multicolumn{3}{c|}{\textbf{Single span}}                                                  & \multicolumn{3}{c|}{\textbf{Multiple spans}}                                                & \multicolumn{3}{c}{\textbf{All spans}}                                                           \\ \cline{3-11} 
                  &                                 & \multicolumn{1}{c|}{\textbf{P}}     & \multicolumn{1}{c|}{\textbf{R}}     & \textbf{F1}    & \multicolumn{1}{c|}{\textbf{P}}     & \multicolumn{1}{c|}{\textbf{R}}     & \textbf{F1}    & \multicolumn{1}{c|}{\textbf{P}}     & \multicolumn{1}{c|}{\textbf{R}}     & \textbf{F1}\\\hline
\parbox[t]{3mm}{\multirow{6.5}{*}{\rotatebox{90}{\rlap{\usebox2}}}}  &   BiLSTM-CRF + Pho2W$_{syllable}$ & \multicolumn{1}{c|}{0.4222}          & \multicolumn{1}{c|}{0.5009}          & 0.4329          & \multicolumn{1}{c|}{0.5134}          & \multicolumn{1}{c|}{0.5712}          & 0.5068          & \multicolumn{1}{c|}{0.7452}          & \multicolumn{1}{c|}{0.7769}          & 0.7453          \\ [1.5pt] \cdashline{2-11} 
                  & XLM-R$_{Base}$                               & \multicolumn{1}{c|}{0.7604} & \multicolumn{1}{c|}{0.7653} & 0.7203 & \multicolumn{1}{c|}{0.7927} & \multicolumn{1}{c|}{0.7574} & 0.7327 & \multicolumn{1}{c|}{0.7766}          & \multicolumn{1}{c|}{0.7574}          & 0.7467          \\ [1.5pt] \cdashline{2-11} 
                  & XLM-R$_{Large}$                               & \multicolumn{1}{c|}{\textbf{0.7577}}          & \multicolumn{1}{c|}{\textbf{0.7679}}          & \textbf{0.7214}          & \multicolumn{1}{c|}{\textbf{0.7829}}          & \multicolumn{1}{c|}{\textbf{0.7569}}          & \textbf{0.7357}          & \multicolumn{1}{c|}{\textbf{0.8071}} & \multicolumn{1}{c|}{\textbf{0.7887}} & \textbf{0.7770} \\ [1.5pt] \hline
\parbox[t]{3mm}{\multirow{6}{*}{\rotatebox{90}{\rlap{\usebox3}}}}  &   BiLSTM-CRF + Pho2W$_{word}$ & \multicolumn{1}{c|}{0.3196}          & \multicolumn{1}{c|}{0.4468}          & 0.3594          & \multicolumn{1}{c|}{0.3533}          & \multicolumn{1}{c|}{0.5001}          & 0.4013          & \multicolumn{1}{c|}{0.6823}          & \multicolumn{1}{c|}{0.7489}          & 0.7036          \\ [1.5pt] \cdashline{2-11} 
                  & PhoBERT$_{Base}$                               & \multicolumn{1}{c|}{0.7392} & \multicolumn{1}{c|}{0.7485} & 0.7016 & \multicolumn{1}{c|}{0.7761} & \multicolumn{1}{c|}{0.7329} & 0.7092 & \multicolumn{1}{c|}{0.7870} & \multicolumn{1}{c|}{0.7680} & 0.7569 \\ [1.5pt] \cdashline{2-11} 
                  & PhoBERT$_{Large}$                               & \multicolumn{1}{c|}{\textbf{0.7435}}          & \multicolumn{1}{c|}{\textbf{0.7567}}          & \textbf{0.7067}          & \multicolumn{1}{c|}{\textbf{0.7878}}          & \multicolumn{1}{c|}{\textbf{0.7557}}          & \textbf{0.7321}          & \multicolumn{1}{c|}{\textbf{0.8028}}          & \multicolumn{1}{c|}{\textbf{0.7835}}          & \textbf{0.7716}          \\ [1.5pt] \hline
\end{tabular}
}
\caption{Experimental results on Single span, Multiple spans, and All spans subsets.}
\label{tab:spanresults}
\end{table*}

% \begin{table}[H] %H
%     \centering
%     \includegraphics[width=1\linewidth]{figures/spans_stats.png} 
%     \caption{Spans quantity and length statistics.}
%     \label{tab:spans_stats}
% \end{table}

\section{Experiments and Results}

\subsection{Baseline Models} 

We treat the task of detecting hate and offensive spans as a task of sequence tagging. As a result, we make use of IOB format \cite{ramshaw-marcus-1995-text} to tag characters for model training, and testing. We conduct experiments on a set of solid baseline models, including BiLSTM-CRF and two pre-trained language models, XLM-R and PhoBERT, to assess the difficulty of our dataset.

%Each token represents a tokenized word.

% \subsection{Models Details}

\textbf{BiLSTM-CRF}: We use BiLSTM-CRF \cite{DBLP:journals/corr/LampleBSKD16}, a model that achieves high performance in the span detection tasks \cite{pavlopoulos2021semeval,nguyen2021span}. We implemented this model with three main layers: (1) The word embedding layer using pre-trained PhoW2V \cite{phow2v_vitext2sql}, (2) The BiLSTM layer, and (3) the Conditional Random Field (CRF).

%\textbf{Transformers}

%Pre-trained language models have recently been proved to have optimal performance in NLP tasks such as NER \cite{Luoma2020exploring}. In this paper, we experiment with several pre-trained language models as below: 

\textit{\textbf{XLM-R:}} XLM-RoBERTa \cite{conneau2019unsupervised} is a multilingual language model and a variant of RoBERTa, pre-trained on 2.5T of data across 100 languages containing 137GB of Vietnamese texts. On several cross-lingual benchmarks, XLM-R outperforms mBERT. 

\textit{\textbf{PhoBERT:}} PhoBERT \cite{nguyen2020phobert} is a monolingual language model which is pre-trained on a 20GB Vietnamese dataset and has the same architecture and approach as RoBERTa. PhoBERT is proven as a state-of-the-art method in multiple Vietnamese-specific NLP tasks such as Part-Of-Speech Tagging, Dependency Parsing, and NER \cite{truong-etal-2021-covid,nguyen2020phobert}.

\subsection{Experimental Settings}
\label{experimental_settings}

\begin{figure*}[h]
\centering
\includegraphics[width=0.7\linewidth, height=4.27cm]{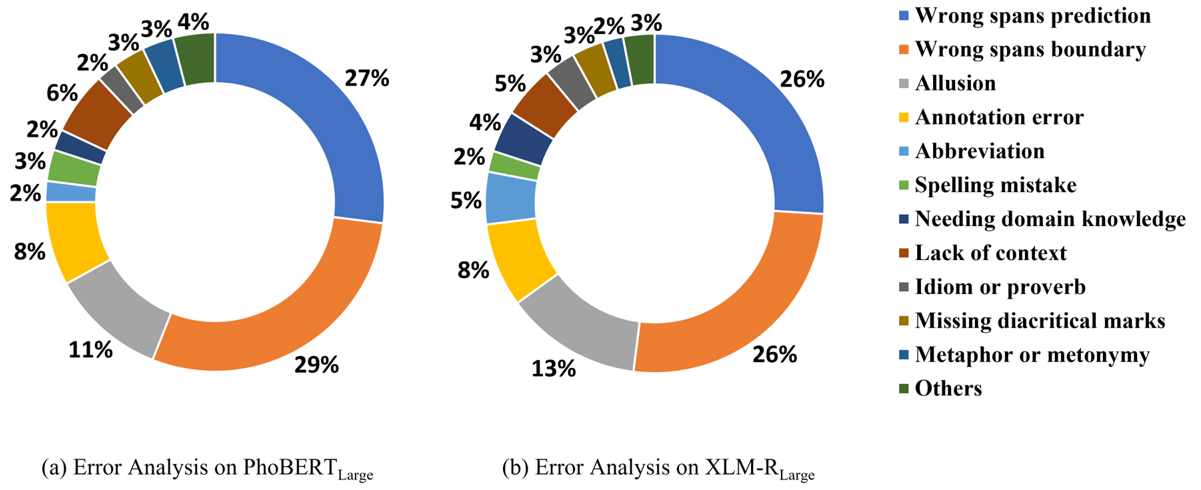} 
\caption{Error analysis conducted on prediction on dev set made by PhoBERT$_{Large}$ and XLM-R$_{Large}$. We divide error cases into 12 categories including wrong spans detection, wrong spans boundary, allusion, annotation error, abbreviation, spelling mistake, needing domain knowledge, lack of context, idiom or proverb, missing diacritical marks, metaphor or metonymy, and others (rare characters, mixing other languages, all words stick together, etc.). These error cases are defined in Appendix \ref{apx:definition_error}.}
\label{fig:err_ana}
\end{figure*}

We empirically fine-tuned all pre-trained language models using \textit{simpletransformers}\footnote{\url{https://simpletransformers.ai/}  (ver.0.63.3)}. For the tokenizer, each comment was tokenized using VnCoreNLP \cite{vu2018vncorenlp} in word-level and syllable-level for fine-tuning the PhoBERT and the XLM-R, respectively. In addition, we used Adam optimizer with a learning rate of 2e-5, a batch size of 8, and trained with 10 epochs.

%Besides, each comment was tokenized using VnCoreNLP (word-level) \cite{vu2018vncorenlp} and syllable-level for fine-tuned PhoBERT and XLM-R, respectively. We implement all pre-trained models using %\textit{simpletransformers}\footnote{\url{https://simpletransformers.ai/}}.
%\textit{simpletransformers}\footnote{\url{https://simpletransformers.ai/} (ver.0.63.3)}.

We utilized a pre-trained word embedding - PhoW2V both syllable-level and word-level settings \cite{phow2v_vitext2sql} with 100 dims to implement the BiLSTM-CRF model. The optimal hyper-parameters of BiLSTM-CRF are described in Table \ref{tab:hyper_params}. All baseline models were trained on a system having 26GB RAM and an NVIDIA Tesla P100 GPU.

\begin{table}[h]
\centering
\begin{tabular}{l||l}
\hline
\textbf{Hyper-parameters} & \textbf{Values} \\ \hline
Optimizer                & Adam           \\ \hdashline
Learning rate            & 0.001          \\ \hdashline
Mini-batch size          & 64             \\ \hdashline
LSTM hidden state size   & 60             \\ \hdashline
Embedding size           & 100            \\ \hdashline
Dropout                  & {[}0.1, 0.1{]} \\ \hdashline
Epochs                   & 10             \\ \hline
\end{tabular}
\caption{Hyper-parameters of the BiLSTM-CRF.}
\label{tab:hyper_params}
\end{table}

\subsection{Evaluation Metrics}
The macro-average F1-score (F1) is used to evaluate our models. For each pair of gold-predicted spans, we compute F1 and then calculate the arithmetic mean of F1 for each of these cases. It should be noted that the final F1-score, Accuracy, and Precision reported are an average of more than ten runs with various random seeds.

\subsection{Experiments and Results}

Table \ref{tab:results} reports the baseline results before and after adding the 5,528 additional clean comments. We discover that after the addition, the performances improve 0.1002$\pm$0.0210. Specifically, PhoBERT$_{Large}$ considerably outperforms other models in the dataset without additional clean data, achieving 0.6867 in F1-score. In addition, the best model trained on Full data is XLM-R$_{Large}$, which has an F1-score of 0.7770. We find that XLM-R$_{Large}$ increased by 0.1014 and PhoBERT$_{Large}$ increased by 0.0849. These results demonstrate that the appearance of the additional clean comments successfully reduces model bias and improves performance.

Table \ref{tab:spanresults} reports our results in three subsets corresponding to Single, Multiple, and All spans. Both Single and Multiple spans subsets are made by the process of splitting the All spans, which also known as the test set, based on the number of spans in each comment. Their results are described as follows:

\begin{table*}
\centering
\begin{tabular}{m{0.5\textwidth}||m{0.2\textwidth}|m{0.2\textwidth}}
\hline
\multirow{2}{*}{\textbf{Ground truth spans }}                                                                                                                                                                                                                                                                                                                         & \multicolumn{2}{c}{\textbf{Models}}                                                     \\ 
\cline{2-3}
                                                                                                                                                                                                                                                                                                                                                          & \multicolumn{1}{c|}{PhoBERT$_{Large}$} & \multicolumn{1}{c}{XLM-R$_{Large}$}  \\ 
\hline
\vspace{1mm}\pbox{0.5\textwidth}{Ns vậy lại xúc phạm cái \hl{đầu b**i} *neutral face emoji*\\(\textbf{Eng}: \textit{If you said so, you insult the \hl{d**k head}~*neutral face emoji*})}\vspace{1mm}                                                                                                                                                                     & {[}"xúc", "đầu"]                            & {[}"b**i"]                                \\ 
\hdashline
\vspace{1mm}\pbox{0.5\textwidth}{@username nhân dân VN tức là cờ đỏ. Còn \hl{đám cờ vàng} là \hl{lũ súc vật lưu vong}. Hiểu hông?\\(\textbf{\textbf{Eng}}: \textit{@username Vietnamese people are red flag. \hl{The yellow flag ones} are \hl{animals and exiled}. Understand?})}\vspace{1mm}                                                                                      & \pbox{0.2\textwidth}{{[}"đỏ.", "đám", "lũ", "súc"]}               & \pbox{0.2\textwidth}{{[}"đám", "lũ", "súc", "vật", "vong."]}   \\ 
\hdashline
\vspace{1mm}\pbox{0.5\textwidth}{\hl{Cướp đêm là giặc, cướp ngày là quan}. 24/7 lúc nào cũng phải nơm nớp. \hl{Than ôi cái đất nước hạnh phúc}\\(\textbf{Eng}:~\textit{\hl{Night thieves~are enemies, day thieves are bureaucrats}. 24/7, we are in a~ perpetual state of fear and anxiety. \hl{What a happy country}})}\vspace{1mm}                     & \pbox{0.2\textwidth}{{[}"Cướp", "cướp", "nớp."]}                  & \pbox{0.2\textwidth}{{[}"Cướp", "giặc,", "cướp"]}              \\ 
\hdashline
\vspace{1mm}\pbox{0.5\textwidth}{Đúng rồi \hl{đéo} thể tin được. Đáng lẽ phải cào bằng ra mới đúng. Thật không thể tin nổi. Phải cào bằng ra mới được\\(\textbf{Eng}: \textit{Yes, \hl{fucking} unbelievable. They should dig until it comes out. Unbelievable. Should dig till it comes out})}\vspace{1mm} & \pbox{0.2\textwidth}{{[}"đéo", "cào", "cào"]}                     & \pbox{0.2\textwidth}{{[}"đéo"] }                                \\
\hline
\end{tabular}
\caption{Case studies in the dev set from ViHOS that are complicated for the PhoBERT$_{Large}$ and XLM-R$_{Large}$. The highlighted spans in first column are Ground truth spans associated with their comments. \textbf{Eng} refers to the English meaning of associated comments.}
\label{tab:exp_err}
\end{table*}

\textbf{Single span}: We experimented with both syllable-level and word-level language models. We discover that the pre-trained language models outperform the BiLSTM-CRF model by 0.3521$\pm$0.0099 in F1. This significant gap proves the fact that word embedding and features extraction of the pre-trained language models on the social media texts are superior to the BiLSTM-CRF. The XLM-R$_{Large}$ model achieves the best performance with a 0.7214 in F1-score. On the other hand, PhoBERT$_{Large}$ achieves a 0.7067 in F1-score. These results show no significant difference in performances among the multilingual and monolingual pre-trained models in the Single span. 

\textbf{Multiple spans}: We experimented with the syllable-level and word-level and found that the pre-trained language models beat the BiLSTM-CRF model by 0.3212$\pm$0.0132. In addition, the performance of XLM-R$_{Large}$ is slightly better than the PhoBERT$_{Large}$ by 0.0036 in F1-score. The results on the Multiple spans are always better than the Single span, which might be explained by the fact that data in Multiple spans comprise more hate and offensive spans that can assist the models in learning more features of the data. 

\textbf{All spans}: The results of the experiments on the All data are higher than the Single span and the Multiple spans. Specifically, in terms of F1-score, results of the XLM-R$_{Large}$ model are higher by 0.0556 and 0.0413 than the highest in the Single span and the Multiple spans, respectively while the figures for the PhoBERT$_{Large}$ are 0.0649 and 0.0395, respectively.

\subsection{Results Analysis}
\label{experiments}
We choose two best models: PhoBERT$_{Large}$ and XLM-R$_{Large}$ to conduct error analysis. As shown in Figure \ref{fig:err_ana}, we report the statistics of the ratio of various types of error cases\footnote{Definition of errors are explained in the Appendix \ref{apx:definition_error}.} of 100 random samples in the dev set. We notice that \textit{wrong spans prediction}\footnote{The model predicts clean spans as hateful and offensive.}, \textit{wrong spans boundary}\footnote{The model predicts inadequate boundary or fails to predict correctly.}, \textit{allusion}\footnote{The comment refers to another person or subject indirectly and disrespectfully.}, \textit{annotation error}\footnote{The annotated span is wrong in terms of our guidelines. There is no later annotation modification in ViHOS.}, and \textit{lack of context}\footnote{The comment has multiple meanings in different contexts, which mislead the prediction.} are major types of prediction failure of the PhoBERT$_{Large}$ and the XLM-R$_{Large}$.

We show some cases from the ViHOS development set in Table \ref{tab:exp_err}. In the first case of "Ns vậy lại xúc phạm cái đầu b**i *neutral face emoji*"  (\textbf{Eng}: \textit{If you said so, you insult the d**k head~*neutral face emoji*}), we notice that the PhoBERT$_{Large}$ could only predict "đầu"$_{head}$ as a hate and offensive span, whereas XLM-R$_{Large}$ predicts "b**i"$_{b**i}$. Both fail to predict the full boundary of "đầu b**i"$_{d**k~head}$. The reason is that asterisks exist in the text ("b**i"). As defined in Subsection \ref{experimental_settings}, the PhoBERT$_{Large}$, which was fine-tuned on the word-level data, cannot identify these tokens, but the XLM-R$_{Large}$, which was fine-tuned on the syllable-level data, can somewhat predict more accurate hate and offensive span even if it has asterisks.

Furthermore, in the second sampled comment, both best models failed to predict "đám cờ vàng" (\textbf{Eng}: \textit{those yellow flag ones}) as a hate and offensive span, owing to the fact that this phase is an offensive metaphor for a rival party to Vietnamese people. In the third instance, the phrase "cướp đêm là giặc, cướp ngày là quan," (\textbf{Eng}: \textit{night thieves are enemies, day thieves are bureaucrats}) which is an idiom that originated from folk poetry, also misleads the prediction. In the final example, the verb "cào"$_{dig}$ has no object and must be comprehended in context. These intriguing and challenging linguistic phenomena encourage more research into more robust models and methods in this field.

\section{Conclusion and Future Work}
\label{conclusion}
We presented ViHOS, a new Vietnamese dataset for evaluating hate and offensive spans detection models. ViHOS includes 26,467 human-annotated spans on 11,056 comments. In addition, state-of-the-art models are conducted as the first baseline models, including BiLSTM-CRF and pre-trained language models such as XLM-R$_{Base}$, XLM-R$_{Large}$, PhoBERT$_{Base}$, and PhoBERT$_{Large}$. As a result, the XLM-R$_{Large}$ model achieves the best performance, with an F1-score of 0.7770. Furthermore, we discover that the performance when detecting multiple spans is better than the performance in detecting single spans in Vietnamese hate and offensive spans detection. Our dataset is available publicly at the GitHub link\footnote{\url{https://github.com/phusroyal/ViHOS}}.

Despite the study's many promising contributions, the proposed research work still has several potential concerns, especially since the performance is still modest, and incorrect predictions could harm users' reputations if they rely heavily on our method. We intend to expand the dataset size and diversity of hate and offensive context for Vietnamese in the future to address this shortcoming. Furthermore, pre-and post-processing techniques will be used to standardize social networking texts \cite{clark2011text} and deal with complex cases (as discussed in Subsection \ref{experiments}) to improve model performance\cite{suman2021astartwice,kotyushev2021mipt,chhablani2021nlrg}, particularly for Vietnamese pre-trained language models. 

\clearpage

\section*{Limitations, Social Impacts, and Ethical Considerations}
\subsection*{Limitations and Social Impacts}

There are numerous incomprehensible comments in our dataset due to the lack of context. Consequently, our annotators had to place themselves in imaginary contexts in order to annotate those comments (see Table \ref{tab:ann_notices} for more details about our solution). This shortcoming combined with the limitations of the neural networks in terms of understanding various linguistic phenomena (see Table \ref{tab:cmt_Characteristic} for more details about nine different linguistic phenomena) caused their performances of this task still insufficient to become practical. 

We also acknowledge the risk associated with publicizing a dataset of hate and offensive spans (e.g. utilizing ours as a source for building abusive chatbots). However, we firmly believe that our proposed benchmark creates more value than risks.

\subsection*{Ethical Considerations}
The undergraduate students in the annotation process are Vietnamese native speakers; have at least 12 years of studying Vietnamese with average scores on the Vietnam National Exam on Literature of 6.5; have at least three years of using social network platforms. They were explicitly warned that their tasks will display hateful and offensive content and if they became overwhelmed, they were also urged to stop labeling. These undergraduate students were paid \$0.1 per comment, which takes an average of 6.44 seconds to complete (excluding the time used by workers who took exceptionally lengthy comments).

All the comments in ViHOS originated in the study of \citealp{luu2021large}, which preserved users' anonymity by removing all of them when creating the ViHSD. As a result, the comments in our dataset do NOT reflect our thoughts or viewpoints. ViHOS is available to the public under a usage agreement for research and related purposes only.
\section*{Acknowledgments}
%We thank all the annotators who help us in building the ViHOS. 
This work has been funded by The VNUHCM-University of Information Technology's Scientific Research Support Fund.
\bibliography{anthology,custom}
\bibliographystyle{acl_natbib}

\appendix

\appendix
\clearpage
\newpage

%\setbox0\hbox{\tabular{c}\textbf{Spans Quantity}\endtabular}
%\setbox1\hbox{\tabular{c}\textbf{Spans Length}\endtabular}
\onecolumn
\section{Abusive Language Characteristics}
\label{apx:Abusive}
% \usepackage{longtable}
% Note: It may be necessary to compile the document several times to get a multi-page table to line up properly
%\begin{longtable}[c]{|m{0.2\textwidth}|m{0.35\textwidth}|m{0.35\textwidth}|}
\begin{longtable}[c]{m{0.2\textwidth}m{0.35\textwidth}m{0.34\textwidth}}
\caption{Characteristics of abusive language in ViHOS with examples, explanations, and solutions for annotators.}
\label{tab:cmt_Characteristic}\\
\hline
\textbf{Abusive language characteristics} &
  \textbf{Examples} &
  \textbf{Explanations and Solutions} \\ \hline
\endfirsthead
\multicolumn{3}{c}%
{{\bfseries Table \thetable\ continued from the previous page.}} \\
\endhead
Non-diacritical marks comments &
  \pbox{0.35\textwidth}{(1) Dit me cai quy trinh,vao cap cuu deo co tien,deo bao hiem thi nam do di. \\ (\textbf{Eng:} \textit{Fuck the procedure, without money or insurance, you could just lay there and no one cares about your emergency}\\ (2) Dung la con linh dien dien vua thoi chang qua nt goi de choc my dien thoi "con ng" nhu linh dien thi ai them\\ (\textbf{Eng:} \textit{That must be the Crazy Linh, so crazy! I just call to tease the Crazy My. No one gonna love Crazy/The person like Crazy Linh!}} &
  \vspace{1mm}
  \pbox{0.34\textwidth}{\textbf{Explanation:} some of these non-diacritical marks comments might trick annotators a little bit. \\ (1): this is a non-diacritical marks comment but still able to understand. \\ (2): there are some problems causing annotators to re-read multiple times as no punctuation, diacritic, and the text "con ng" could be considered as "crazy girl" or "the type (of human)" and both of these meanings is inappropriate.\\ \\ \textbf{Solution:} Non-diacritical marks comments are annotated as others. Annotators have to re-read until they fully understand the texts if needed. Those examples are annotated as follows:\\ (1): {[}"Dit me",  "deo", "deo"{]}\\ (\textbf{Eng:}  {[}"\textit{Fuck}",  "\textit{fuck}", "\textit{fuck}"{]})\\ (2): {[}"con", "dien", "dien", "con ng",  "dien", "ai them"{]}\\ (\textbf{Eng:}  {[}"\textit{con}",  "\textit{crazy}", "\textit{crazy}", "\textit{crazy/the person like}",  "\textit{crazy}", "\textit{No one gonna love}"{]} in which the word "con" is an inappropriate way to call a woman.)}\vspace{1mm} \\ \hline
Metaphors, metonymies &
  \pbox{0.35\textwidth}{(1) cái miệng rộng quá đẻ con ra còn lọt\\ (\textbf{Eng:} \textit{The mouth is too big that a baby could even be born through it})\\ (2) Dm! qua đợt dịch này thì thằng sống ích kỷ này chắc sớm gia nhập juventuts\\ (\textbf{Eng:} \textit{Fuck it! After the pandemic, this selfish boy will soon join Juventuts})}                                                     &
  \vspace{1mm}
  \pbox{0.34\textwidth}{\textbf{Explanation:} in our dataset, many comments use metaphors or metonymy to convey their hate or offensiveness in another way. \\ (1): this is a metaphor of a mouth as a vagina.\\ (2): this is a metonymy of Juventus' jersey as prison shirts (a common metonymy in Vietnamese).\\ \\ \textbf{Solution:} Annotators are required to annotate the whole ideas of metaphors, metonymy.\\ (1):  {[}"cái miệng rộng quá đẻ con ra còn lọt"{]}\\ (2):  {[}"Dm", "thằng sống ích kỷ này chắc sớm gia nhập juventuts"{]}\\ (\textbf{Eng:}  {[}"\textit{Fuck}",  "\textit{this selfish boy will soon join Juventuts}"{]}) \vspace{1mm}} \\ \hline
Puns &
  \pbox{0.35\textwidth}{(1) Ad đăng bài này cũng là Bồn Kỳ Lắc nè nè nè :) \\ (\textbf{Eng:} \textit{The admin, who posts this, is also a "Bồn Kỳ Lắc")}} &
  \vspace{1mm}\pbox{0.34\textwidth}{\textbf{Explanation:} Some comments use phrases that only read them backwards, they make sense. As in the example, "Bồn Kỳ Lắc", if this phrase is read backwards, it is "Bắc Kỳ Lồn" (pussy north).\\ \\ \textbf{Solution:} Annotators are required to annotate these puns too.\\ (1):  {[}"Bồn Kỳ Lắc"{]}}\vspace{1mm} \\ \hline
Using non-words characters to form hieroglyphs &
  \pbox{0.34\textwidth}{(1) có tin t lấy *knife symbol* xiên chết cụ m ko :)))\\ (\textbf{Eng:} \textit{Do you belive that I could get a *knife symbol* to fucking kill you?})\\ (2) Đâm vào () \\ (\textbf{Eng:} \textit{stab in the ()}} &
  \vspace{1mm}\pbox{0.34\textwidth}{\textbf{Explanation:} \\ (1): the *knife symbol* is used instead of the word knife. \\ (2): the existence of "()" can be considered as pussy in this context.\\ \\ \textbf{Solution:} Annotating the non-words only if they can convey a full meaning of hate or offensiveness, and the whole phase if they can not.\\ (1): {[}"t", "*knife symbol*", "xiên chết cụ m""{]}\\ (\textbf{Eng:}  {[}"\textit{t}",  "\textit{*knife symbol*}", "\textit{fucking kill you}"{]}) in which "t", "m" are inappropriate pronouns.)\\ (2): {[}"Đâm vào ()"{]}}\vspace{1mm} \\ \hline
Spelling mistakes &
  \pbox{0.35\textwidth}{(1) Tôi đeo hiểu bạn noi cai gì luôn a :))) \\ (\textbf{Eng:} \textit{I have no fucking idea about what you saying})\\ (2) Mà dx cái chữi ngta nghe hài vcl bù lại củng đở  \\ (\textbf{Eng:} \textit{Those fucking curses is so funny that can even refill that})\\(3)Khi bạn xai trính tã nhưng cá ghén vít đún trính tã :))\\ (\textbf{Eng:} \textit{When you make spelling mistakes but trying to fix it :))})} &
  \vspace{1mm}\pbox{0.34\textwidth}{\textbf{Explanation:} \\ (1): the words "đeo," "noi," "cai," "a" are spelling mistakes. The words can be understood as "đéo," "nói," "á". As that, these words are the same in accidentally missing acute marks. \\ (2): the words: "chữi," "củng," "đở" are spelling mistakes is a phenomenon of mistaking tilde mark for hook above mark, and this often happens in some parts of Vietnam \cite{nguyenhoainguyen}. There are also familiar phenomena of mistaking marks such as tilde mark for underdot mark, acute mark for hook above mark, and so on \cite{nguyenhoainguyen}. \\ (3): the phases: "xai trính tã," "cá ghén," "vít đúng trính tã" are spelling mistakes but on purpose. This comment utilizes spelling mistakes to attack opponents who also have spelling mistakes. Furthermore, these spelling mistakes also abuse opponents based on regional distinctions in accent, which cause some phenomenon of mistaking diacritical marks as in Example (2).\\ \\ \textbf{Solution:} The same as dealing with non-diacritical marks comments, annotators work as usual.\\ (1): {[}"deo"{]}\\ (\textbf{Eng:}  {[}"\textit{fuck}"{]})\\ (2): {[}"chữi", "vcl"{]}\\ (\textbf{Eng:}  {[}"\textit{curse}", "\textit{fuck}"{]})\\ (3): {[}"bạn xai trính tã nhưng cá ghén vít đún trính tã :))"{]}}\vspace{1mm} \\ \hline
Allusive language &
  \vspace{1mm}\pbox{0.35\textwidth}{(1) ra gì thì toi rồi, nó khác gì cách đảng CS chọn người, hồng hơn chuyên. Suy nghĩ kĩ đi. Giải độc cộng sản đã khó, giải độc tư tưởng cánh tả còn khó hơn.\\ (\textbf{Eng:} \textit{If it had something, it was done! It is just like the way CS (stands for Communism) chooses people, beauty over the profession. Detoxifying Communism is hard; detoxifying the left-wing political ideologies is even harder.})\\ (2) Rút kinh nghiệm lại được xài nghìn tỷ\\ (\textbf{Eng:} \textit{Just say learned and then can use trillion VND})}\vspace{1mm} &
  \pbox{0.34\textwidth}{\textbf{Solution:} Annotators are required to annotate the whole profound abuse.\\ (1): {[}"toi", "hồng hơn chuyên", "Giải độc cộng sản đã khó, giải độc tư tưởng cánh tả còn khó hơn"{]}\\ (\textbf{Eng:}  {[}"\textit{die}", "\textit{beauty over the profession}", "\textit{Detoxifying Communism is hard; detoxify the left-wing political ideologies even harder}"{]})\\ (2): {[}"Rút kinh nghiệm lại được xài nghìn tỷ"{]}} \\ \hline
Homonym &
  \vspace{1mm}\pbox{0.35\textwidth}{(1) coin card \\ ("coin card" is a homonym of "con cặc", which means \textit{dick})}\vspace{1mm} &
  \vspace{1mm}\pbox{0.34\textwidth}{\textbf{Solution:} annotate the whole abusive homonym.\\ (1): {[}"coin card"{]}}\vspace{1mm} \\ \hline
Mixing languages &
  \vspace{1mm}\pbox{0.35\textwidth}{(1) Vl fake\\ \textbf{Eng:} Fuck! Fake)\\ (2) phe \textit{X} compat tổng \textit{Y} =))\\ (\textbf{Eng:} \textit{\textit{X}'s side combats \textit{Y}'s side})\\ (3) Tối Thầy stream đá phò đi thầy ơi\\ (\textbf{Eng:} \textit{You should stream fucking some whores tonight, please!})}\vspace{1mm} &
  \vspace{1mm}\pbox{0.34\textwidth}{\textbf{Solution:} treat foreign words as others and annotate if they meet the definition of hate and offensive spans.\\ (1): {[}"Vl", "fake"{]}\\ (\textbf{Eng:}  {[}"\textit{fuck}", "\textit{fake}"{]})\\ (2): {[}"compat"{]}\\ (\textbf{Eng:}  {[}"\textit{combat}"{]})\\ (3): {[}"đá phò"{]}\\ (\textbf{Eng:}  {[}"\textit{fuck some whores}"{]})}\vspace{1mm} \\ \hline
Trick hate speech detecting systems on purpose &
  \pbox{0.35\textwidth}{(1) Đấy Ông già xạo l*n của các bạn được tắm bồn Thái Lan đấy=)))\\ (\textbf{Eng:} \textit{See, your fucking old liar is Thai bathing again =)))})\\ (2) thằng  c h.ó ngu\\ (\textbf{Eng:} \textit{Fucking stupid guy})\\ (3) vualonemlabaonhieu cm\\ (\textbf{Eng:} \textit{How long in cm to fit your pussy?})} &
  \vspace{1mm}\pbox{0.34\textwidth}{\textbf{Explanation}: some hate or offensive comments use punctuation to censor their inappropriate words (as in (1), asterisk is used in the word l*n (pussy)) or to disunite characters in words (as in (2), dot and space are used to disunite the words chó (dog) into c h.ó). \\ These efforts actually can trick many hate speech detecting systems, or to put all words together (as in (3), vualonemlabaonhieu cm should be "vừa lồn em là bao nhiêu cm").\\ \\ \textbf{Solution:} we annotate all characters which can be form into hate or offensive phase.\\ (1): {[}"xạo l*n"{]}\\ (\textbf{Eng:}  {[}"\textit{fucking lie}"{]})\\ (2): {[}"thằng  c h.ó", "ngu"{]}\\ (3): {[}"lon"{]}\\ (\textbf{Eng:}  {[}"\textit{pussy}"{]})}\vspace{1mm} \\ \hline
\end{longtable}
\section{Notices for annotators}
\label{apx:notiecs}
% \clearpage
% \newpage
\begin{longtable}[c]{m{0.2\textwidth}m{0.35\textwidth}m{0.35\textwidth}}
%\begin{longtable}{p{4cm}p{4cm}p{6cm}}
\caption[Additional notices for annotators.]{Additional notices for annotators.}
\label{tab:ann_notices}\\
\hline
\textbf{Notices} &
  \textbf{Example} &
  \textbf{Explanation} \\ \hline
\endfirsthead
\multicolumn{3}{c}%
{{\bfseries Table \thetable\ continued from the previous page.}} \\
\endhead
Try to figure out and consider as in the original context of the comments. &
   &
  This could help annotators understand complex and non-context hate, offensive comments. \\ \hline
Do not let emotion affect the annotating process. &
   &
  Annotators exposed in a long time to toxic comments are reported to get used to the frequently appearing hate, offensiveness. \\ \hline
Check the provided Vietnamese Dictionary if there is any uncertainty in being sure a word is hate or offensive. &
   &
  We use the most reputable Vietnamese dictionary \cite{hoangphe} to provide to annotators. \\ \hline
Should span the object is compared to in an inappropriate comparison. &
  \pbox{0.35\textwidth}{"Ăn cơm nhìn như chó"\\ (\textbf{Eng:} \textit{You eat like a dog}) \\ \textbf{Spans:} {[}"nhìn như chó"{]}} &
  We consider this is an inappropriate comparison in which annotators must span "chó"$_{dog}$. \\
  \hline
However, we should span the whole comparison if spanning only the object is compared to might not convey the complete hate or offensive idea &
  \pbox{0.35\textwidth}{"ý thức như trẻ lớp mầm" \\ \textbf{Spans:} {[}"ý thức như trẻ lớp mầm"{]}\\ (\textbf{Eng:} \textit{Your awareness is just like a kindergarten kid})} &
  If we only span "trẻ lớp mầm" (kindergarten kid), it will not convey the complete offensive idea. As that, annotators are encouraged to span the whole text instead. \\ \hline
Do not span conjunctions; exceptional Vietnamese cases; standard ways to call LGBTQ+. &
  \pbox{0.35\textwidth}{(1) Vì thế, nên, nhưng, mà, etc.\\ (\textbf{Eng:} \textit{so, so, but, but, etc.}) \\  (2) Gay, les, etc\\ (3) Bóng, bê đê, ái nam ái nữ, etc.} &
  \vspace{1mm}\pbox{0.35\textwidth}{(1) Some conjunctions in Vietnamese.\\ (2) Appropriate ways to specify LGBTQ+.\\ (3) Inappropriate ways to specify LGBTQ+ that need to be spanned in comments.}\vspace{1mm} \\ \hline
Blatant hate and offensive words prioritize being spanned over the others, especially in sentences without diacritics and could not be understanded. &
  \pbox{0.35\textwidth}{May Cha H O an Tro. Cap. Ranh c. Di ngoi le. Nhieu chuyen. Do  Cai Thu ,  do tam than\\ \textbf{Spans:} {[}"Cap", "ngoi le", "Nhieu chuyen", "do tam than"{]}\\ (\textbf{Eng:} {[}"\textit{steal}", "\textit{gossip}", "\textit{talkative}", "\textit{psycho}"{]})} &
  Similar to this non-diacritical and incomprehensible comment, words as highlighted ones are more straightforward to be spanned. \\ \hline
Span the whole phase violating human rights. &
  \pbox{0.35\textwidth}{"Về nước anh nên vào tù ở trước thay vì đi cách ly." \\ (\textbf{Eng:} \textit{When you return to Vietnam, you should go to jail first instead of going to isolation}) \\ \textbf{Spans:} {[}"Về nước anh nên vào tù ở trước thay vì đi cách ly"{]}} &
  Comments violate human rights, usually complex to specify hate, offensive words to span. As in this example, a Vietnamese citizen comes back from a foreign country has a right to have isolated healthcare firstly. \\ \hline
Span the whole obscene acronyms. &
  \vspace{1mm}\pbox{0.35\textwidth}{(1) clgv \\ (\textbf{Eng:} \textit{wtf is that?}) \\ (2) clmn \\ (\textbf{Eng:} \textit{your mom’s pussy}) \\ (3) cmn \\ (\textbf{Eng:} \textit{your mother}).} \vspace{1mm} &
   \\ \hline
Follow the rule with words that do not have diacritics and conjoin to span out the hate, offensive spans. &
  \pbox{0.35\textwidth}{cailongithe \\ (\textbf{Eng:} \textit{wtf is that?}) }&
  Some comments have strings being constructed by many words missing diacritical marks, but still able to understand. The annotators should only span the hate and offensive characters set out of the string as in the example. \\ \hline
Span hate, offensiveness separately. &
  \vspace{1mm}\pbox{0.35\textwidth}{"Cả đám dlv là nhóm ngu dốt, trẻ trâu potay vcl luôn." \\ (\textbf{Eng:} \textit{The whole dlv crew are stupid, bull-headed kids (so I have) no fucking thing to say.})\\ \textbf{Spans:} {[}"đám", "ngu dốt", "trẻ trâu"{]}\\ (\textbf{Eng:} {[}"\textit{crew}", "\textit{stupid}", "\textit{bull-headed kids}"{]})}\vspace{1mm} &
  This comment must be spanned as in the example, but not as "đám," "ngu dốt, trẻ trâu," "vcl," "NGU." \\ \hline
\end{longtable}
% \end{landscape}

%\section{Spans Quantity and Length Statistics.}
%\label{app:LengthStatistics}

\twocolumn
\section{Definition of Error Cases for Error Analysis}
\label{apx:definition_error}
We introduce 12 error definitions as follows:

\begin{enumerate}
    \item \textbf{Wrong spans prediction}: The model predicts clean spans as hateful and offensive.
    
    \item \textbf{Wrong spans boundary}: The model predicts inadequate boundary or fails to predict correctly.
    
    \item \textbf{Allusion}: The comment refers to another person or subject in an indirect and disrespectful way.

    \item \textbf{Annotation error}: Annotators have improperly annotated the span. The reason might be that they somehow do not follow the provided guidelines. However, there is no modification in the final dataset.

    \item \textbf{Abbreviation}: The comment contain short forms of words.

    \item \textbf{Spelling mistake}: The comment is spelling mistake.

    \item \textbf{Needing domain knowledge}: Dialect and professional expertise are required to detect span in comments.

    \item \textbf{Lack of context}: In different contexts, the comment could be understand in multiple meaning.

    \item \textbf{Idiom or proverb}: The comment contains idiom or proverb.

    \item \textbf{Missing diacritical marks}: Words in the comment do not have diacritical marks.

    \item \textbf{Metaphor or metonymy}: The comment contains metaphor or metonymy.

    \item \textbf{Others}: The comment contains rare characters, other languages, words in it are stick together, etc.

\end{enumerate}

\end{document}